%% file: aaai25.tex
\documentclass[letterpaper]{article} 

\usepackage{aaai25}

\usepackage{times}  
\usepackage{helvet}  
\usepackage{courier}  
\usepackage[hyphens]{url}  
\usepackage{graphicx} 
\usepackage{natbib}  
\usepackage{caption} 
\usepackage{algorithm}
\usepackage{algorithmic}

\usepackage{newfloat}
\usepackage{listings}
\DeclareCaptionStyle{ruled}{labelfont=normalfont,labelsep=colon,strut=off} 
\floatstyle{ruled}
\newfloat{listing}{tb}{lst}{}
\floatname{listing}{Listing}
\usepackage{amsmath}
\usepackage{amssymb}
\usepackage{amsbsy} 

\usepackage{enumitem}
\usepackage{booktabs}
\usepackage{subfigure}

\usepackage{xcolor}

\title{Tensor Completion for Surrogate Modeling of Material Property Prediction}
\author {
    Shaan Pakala,
    Dawon Ahn,
    Evangelos Papalexakis
}
\affiliations {
    UC Riverside\\
    spaka002@ucr.edu, 
    dahn017@ucr.edu, 
    epapalex@cs.ucr.edu
}

\begin{document}

\maketitle

\input{000abstract}
\input{010introduction}

\input{020methods}

\input{030experiments}
\input{040conclusion}

\section*{Acknowledgements}
\small{Research was supported by the National Science Foundation CAREER grant no. IIS 2046086, grant no. IIS 1901379, and CREST Center for Multidisciplinary Research Excellence in Cyber-Physical Infrastructure Systems (MECIS) grant no. 2112650.}

\bibliography{aaai25}

\end{document}

%% file: 000abstract.tex
\begin{abstract}

When designing materials to optimize certain properties, there are often many possible configurations of designs that need to be explored. For example, the materials' composition of elements will affect properties such as strength or conductivity, which are necessary to know when developing new materials. Exploring all combinations of elements to find optimal materials becomes very time consuming, especially when there are more design variables. For this reason, there is growing interest in using machine learning (ML) to predict a material's properties. In this work, we model the optimization of certain material properties as a tensor completion problem, to leverage the structure of our datasets and navigate the vast number of combinations of material configurations. Across a variety of material property prediction tasks, our experiments show tensor completion methods achieving decreased error in 2 of our 4 material property prediction tasks, compared with baseline ML models such as GradientBoosting and Multilayer Perceptron (MLP), maintaining similar training speed.

\end{abstract}

%% file: 010introduction.tex
\section{Introduction}

Material property prediction is a crucial aspect of designing new materials, in order to design a material with optimal properties. This includes designing strong materials \cite{gongora2024accelerating}, magnetic materials \cite{wang2020accelerated}, and semiconductor materials \cite{haghshenas2024full}. Machine learning (ML) is becoming a very prominent technique for this problem, and we see both composition-based (using the chemical formula) and structure-based (using the structure) material property prediction \cite{gongora2024accelerating,hu2024realistic,li2024md}.

Among the material properties being predicted, there is a lot of attention around predicting band gap, which directly affects the conductivity of materials \cite{alsalman2023bandgap,hu2024realistic}. This is very useful in fields such as semiconductor development, where knowing the conductivity is crucial. There is also work on predicting the formation energy \cite{jha2019enhancing,xie2018crystal,jha2022moving}. We also see prediction of magnetism, which is important for designing materials with ideal magnetic properties \cite{li2022machine}.

Unfortunately, issues tend to arise with a lack of training data \cite{xin2021active}, as is typical in any ML problem, especially when deep learning is introduced. In this work, we use tensor completion algorithms, designed for sparse datasets, to infer a material's properties \cite{sidiropoulos2016tensor}.

We model material property prediction tasks as tensors. This allows us to use tensor completion for composition-based material property prediction, where we use the chemical formula to predict material properties. By leveraging tensor completion algorithms, we take advantage of the structure of our datasets to infer material property values.

%% file: 020methods.tex
\section{Methods}

Here we outline our methods for dataset generation and tensor completion for materials' property prediction.

\subsection{Tensor Dataset Generation}

To generate tensors used for our experiments, each tensor mode corresponds to either a unique element in the material, or the amount of each unique element in the material. An example fourth order tensor is shown below. The highlighted value represents the chemical formula AuBr$_{5}$.

\begin{figure}[H]
    \centering
    \begin{center}
        \includegraphics[width = 0.41\textwidth]{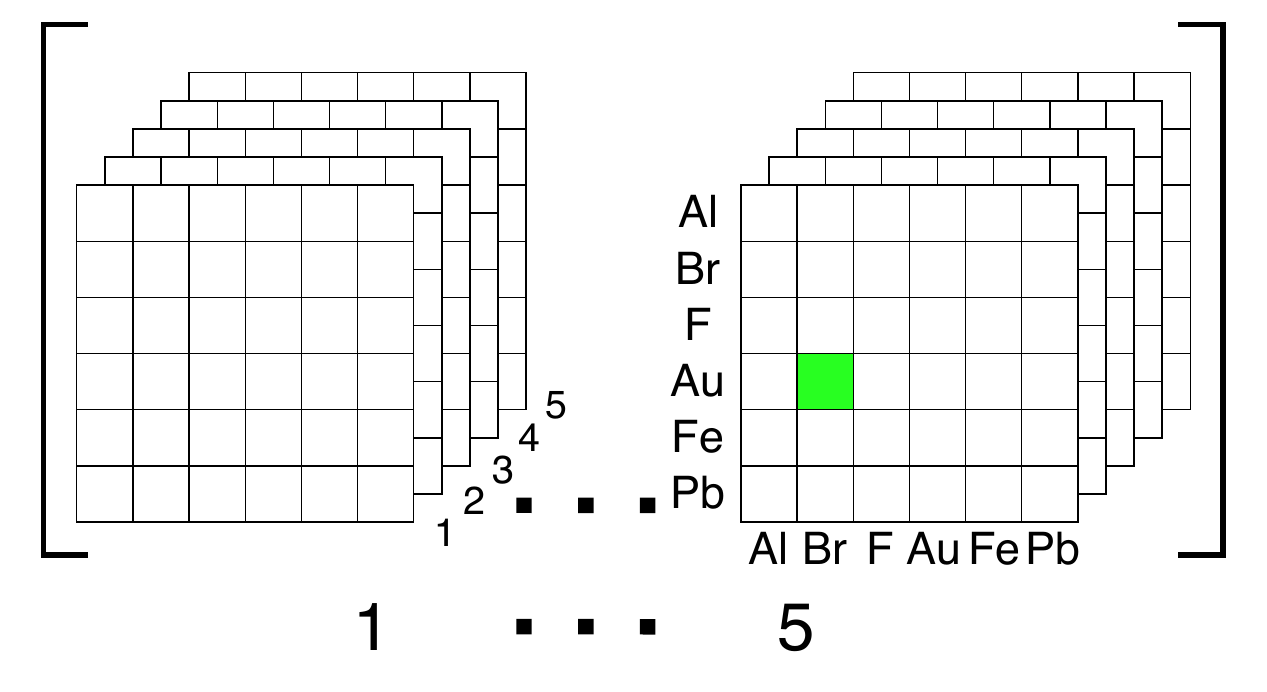}
        \caption{Example fourth order tensor representing materials' elements and elements' ratios. The highlighted entry could correspond to the chemical formula AuBr$_{5}$, since it corresponds to indices Au \& Br for the elements, and indices 1 \& 5 for the number of atoms in the material.}
    \end{center}
\end{figure}

\subsection{Tensor Completion for Materials Design}

By using tensor completion, we leverage the inherent structure in our multidimensional datasets to predict the results of the entire space of combinations, using few training values.

\subsubsection{Tensor Decomposition}
Tensor decomposition is a key technique to extract information from tensor data. For example, CPD (Canonical Polyadic Decomposition) \cite{PARAFAC} is a common approach, factoring the tensor into the sum of rank 1 tensors. Tensor decomposition is crucial for tensor completion, which fills in a tensor's missing values. 

\subsubsection{Tensor Completion Models} 
In our experiments, we use a variety of tensor completion models. These include \textbf{CPD} \cite{PARAFAC}, \textbf{CPD-S} \cite{10825934, ahn2021time}, and \textbf{NeAT} \cite{ahn2024neural}.

\subsubsection{Performance Evaluation}

We randomly sample train values, using the rest to calculate Mean Absolute Error (MAE). For true value $y$ and predicted $\hat y$, $MAE = \frac{1}{n}\sum_{i=1}^{n}|y_i - \hat y_i|$.

%% file: 030experiments.tex
\section{Experimental Evaluation}

Here we experimentally evaluate the viability of using tensor completion on the following tasks that we have defined:

\begin{enumerate}[label=\textbf{Task \arabic*:}, leftmargin=1.3cm]
    \item Total magnetization prediction, using chemical formula. 100,000 training, 50,000 test values. \footnotemark[1]
    \item Formation energy prediction, using chemical formula. 45,000 training, 4213 test values. \footnotemark[2]
    \item Band gap prediction, using chemical formula. 45,000 training, 4213 test values.\footnotemark[2]
    \item Band gap prediction, using chemical formula. 1500 training, 759 test values. \footnotemark[3]
\end{enumerate}

\footnotetext[1]{Original data from \cite{lematerial_2024}.}
\footnotetext[2]{Original data from \cite{li2024md}.}
\footnotetext[3]{Original data from \cite{hu2024realistic}.}

\subsection{Tensor Model Comparison*}

\renewcommand{\thefootnote}{*}
\footnotetext{There was misleading results due to highly redundant datasets, these results have now been updated. Table 1 contains the old results with redundant data, Table 2 contains the new results with the redundancies removed.}
\renewcommand{\thefootnote}{\arabic{footnote}}

For each tensor completion model, we display the average MAE over 5 iterations. We also compare non-tensor models: GradientBoosting \cite{scikit-learn}, XGBoost \cite{Chen_2016}, and an MLP \cite{paszke2019pytorch}.

\begin{table}[H]
    \centering
    \begin{tabular}{lllll}
    \toprule
     & Task 1 & Task 2 & Task 3 & Task 4 \\
    \midrule
    CPD & 1.397 & 0.570 & 0.650 & 0.430 \\
    CPD-S & 1.428 & 0.627 & 0.631 & \textbf{0.390} \\
    NeAT & 1.323 & 0.564 & \textbf{0.608} & 0.420 \\
    \midrule
    Non-Tensor Models\\
    \midrule
    GradientBoosting & \textbf{0.898} & \textbf{0.307} & 0.820 & 0.560 \\
    XGBoost & 1.084 & 0.671 & 0.686 & 0.462 \\
    MLP & 1.242 & 0.586 & 0.746 & 0.574 \\
    \bottomrule
    \end{tabular}
    
    \caption{Comparison of ML models, both tensor and non-tensor methods, using the MAE over 5 iterations. \label{tasks_table}}
    
\end{table}

\begin{table}[H]
    \centering
    \begin{tabular}{lllll}
    \toprule
     & Task 1 & Task 2 & Task 3 & Task 4 \\
    \midrule
    CPD & 1.179 & 0.344 & 0.467 & 0.428 \\
    CPD-S & 1.254 & 0.369 & 0.487 & \textbf{0.396} \\
    NeAT & \textbf{0.907} & \textbf{0.277} & \textbf{0.456} & 0.426 \\
    \hline
    Non-Tensor Models & & & & \\
    \hline
    HGB & 1.018 & 0.280 & 0.774 & 0.560 \\
    XGB & 1.212 & 0.643 & 0.618 & 0.454 \\
    MLP & 1.403 & 0.677 & 0.709 & 0.567 \\
    \bottomrule
    \end{tabular}
    
    \caption{Same as Table \ref{tasks_table}, but we removed the redundancy in our datasets. Comparison of ML models, both tensor and non-tensor methods, using the MAE over 5 iterations. \label{tasks_table_updated}}
    
\end{table}

The table shows tensor completion regularly outperforms the benchmark non-tensor models across a variety of tasks.

\subsection{Individual Samples' Comparison}

To observe how useful our methods are in practice, we look at 5 randomly sampled values and predictions for each task.

\begin{figure}[H]
    \centering
    \begin{center}
        \includegraphics[width = 0.42\textwidth]{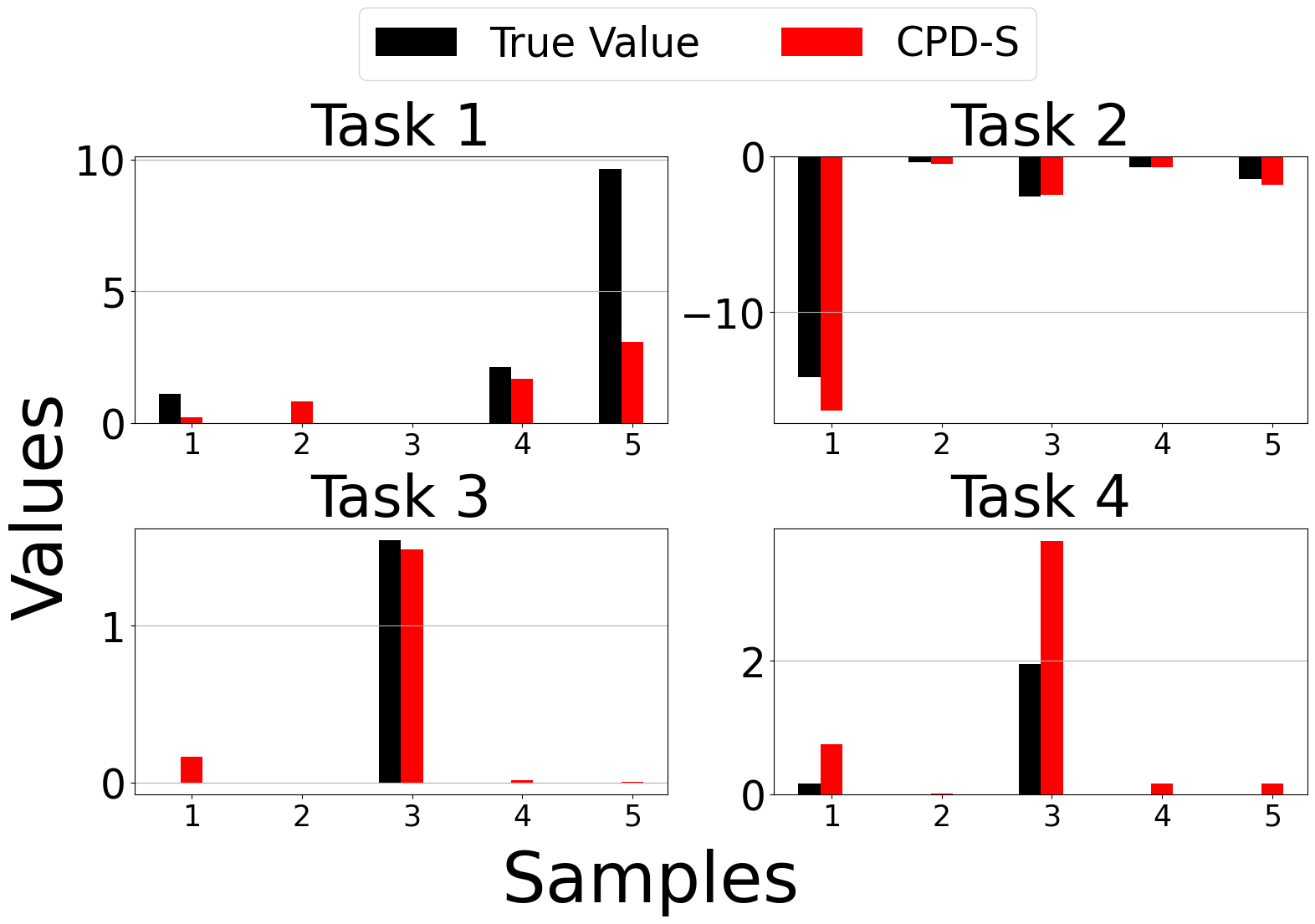}
        \caption{Random samples' predicted and actual values. Overall CPD-S offers reliability. Even with limited training samples in Task 4, CPD-S is still able to distinguish between low and high band gap values.}
    \end{center}
\end{figure}

\subsection{Efficiency Analysis}

Here we observe the performance and training time for different models with respect to the number of training values.

\begin{figure}[!ht]

    \centering
    \subfigure[MAE]{
        \includegraphics[width=0.45\linewidth, height = 2.75cm]{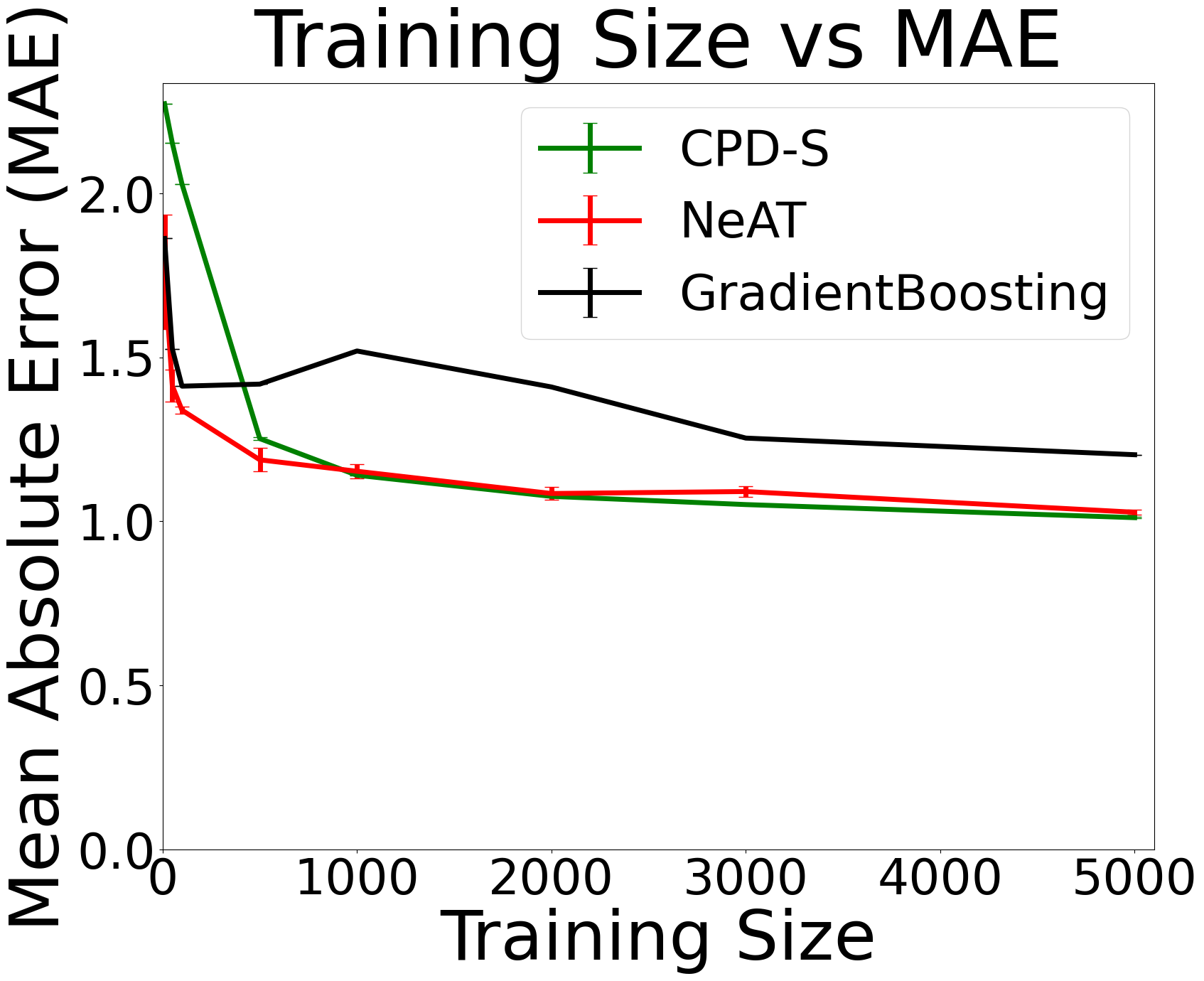}
    }
    \subfigure[Training Time]{
        \includegraphics[width=0.45\linewidth, height = 2.75cm]{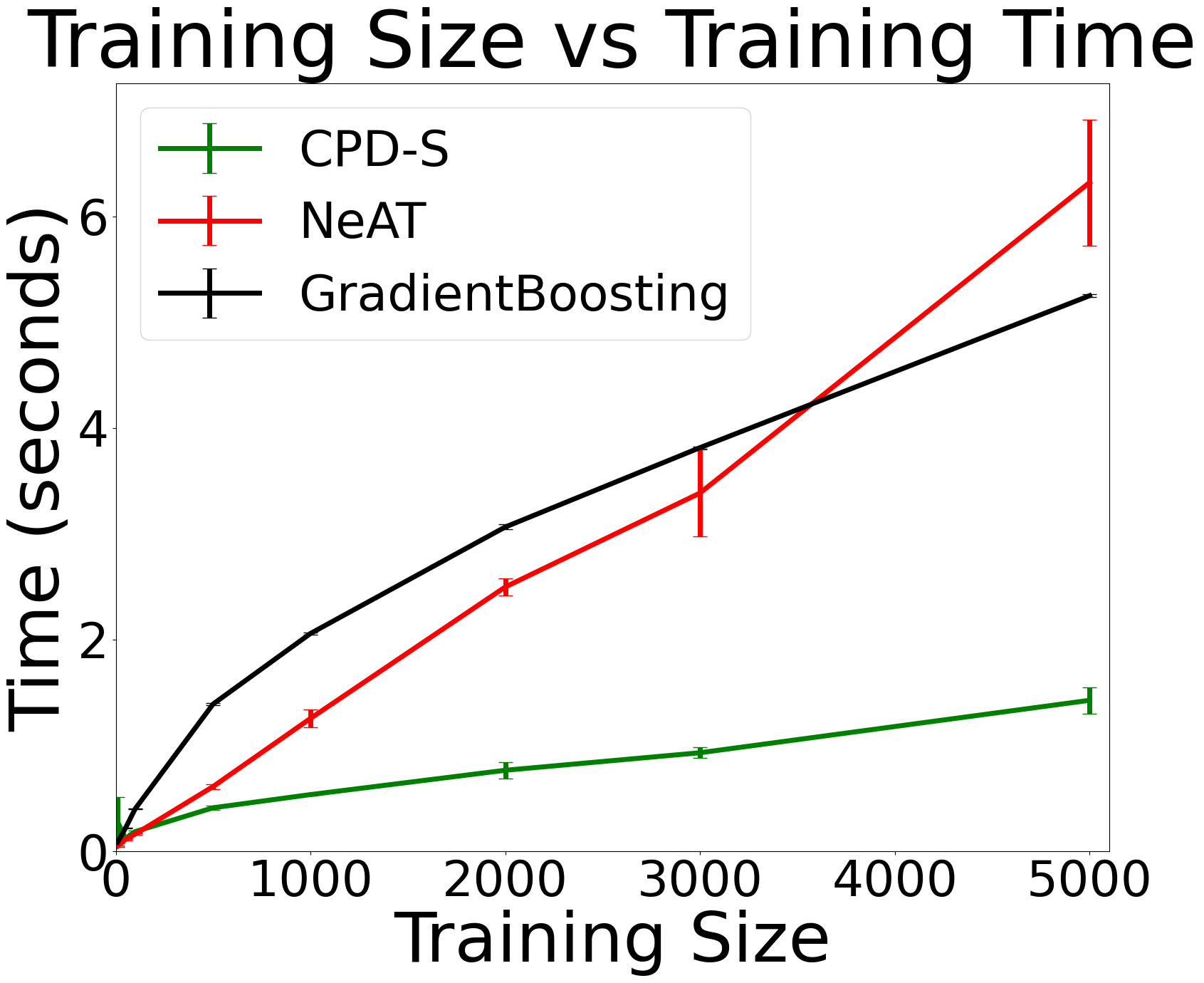} 
    } 

    \caption{MAE and train time for models, with respect to number of train entries. This displays for a range of training sizes, CPD-S performs well without requiring more time.}

\end{figure}

%% file: 040conclusion.tex
\section{Conclusion}

By modeling material property prediction tasks as tensor completion problems, we leverage the inherent structure in our datasets to fill in missing entries. We show the viability of modeling various types of composition-based material property prediction problems as tensor completion problems, predicting band gap, formation energy, and total magnetization. This way we significantly accelerate material design problems by using just a small fraction of computed values to infer the results unobserved materials' values.

%% file: aaai25.bbl
\begin{thebibliography}{20}
\providecommand{\natexlab}[1]{#1}

\bibitem[{Ahn, Jang, and Kang(2021)}]{ahn2021time}
Ahn, D.; Jang, J.-G.; and Kang, U. 2021.
\newblock Time-aware tensor decomposition for sparse tensors.
\newblock In \emph{2021 IEEE 8th International Conference on Data Science and Advanced Analytics (DSAA)}, 1--2. IEEE.

\bibitem[{Ahn et~al.(2024)Ahn, Saini, Papalexakis, and Payani}]{ahn2024neural}
Ahn, D.; Saini, U.~S.; Papalexakis, E.~E.; and Payani, A. 2024.
\newblock Neural Additive Tensor Decomposition for Sparse Tensors.
\newblock In \emph{Proceedings of the 33rd ACM International Conference on Information and Knowledge Management}, 14--23.

\bibitem[{Alsalman, Alqahtani, and Alharbi(2023)}]{alsalman2023bandgap}
Alsalman, M.; Alqahtani, S.~M.; and Alharbi, F.~H. 2023.
\newblock Bandgap energy prediction of senary zincblende III--V semiconductor compounds using machine learning.
\newblock \emph{Materials Science in Semiconductor Processing}, 161: 107461.

\bibitem[{Chen and Guestrin(2016)}]{Chen_2016}
Chen, T.; and Guestrin, C. 2016.
\newblock XGBoost: A Scalable Tree Boosting System.
\newblock In \emph{Proceedings of the 22nd ACM SIGKDD International Conference on Knowledge Discovery and Data Mining}, KDD ’16, 785–794. ACM.

\bibitem[{Gongora et~al.(2024)Gongora, Friedman, Newton, Yee, Doorenbos, Giera, Duoss, Han, Sullivan, and Rodriguez}]{gongora2024accelerating}
Gongora, A.~E.; Friedman, C.; Newton, D.~K.; Yee, T.~D.; Doorenbos, Z.; Giera, B.; Duoss, E.~B.; Han, T. Y.-J.; Sullivan, K.; and Rodriguez, J.~N. 2024.
\newblock Accelerating the design of lattice structures using machine learning.
\newblock \emph{Scientific Reports}, 14(1): 13703.

\bibitem[{Haghshenas et~al.(2024)Haghshenas, Wong, Sethu, Amal, Kumar, and Teoh}]{haghshenas2024full}
Haghshenas, Y.; Wong, W.~P.; Sethu, V.; Amal, R.; Kumar, P.~V.; and Teoh, W.~Y. 2024.
\newblock Full prediction of band potentials in semiconductor materials.
\newblock \emph{Materials Today Physics}, 46: 101519.

\bibitem[{Harshman(1970)}]{PARAFAC}
Harshman, R. 1970.
\newblock Foundations of the PARAFAC procedure: Models and conditions for an" explanatory" multimodal factor analysis.
\newblock \emph{UCLA working papers in phonetics}.

\bibitem[{Hu et~al.(2024)Hu, Liu, Fu, and Dong}]{hu2024realistic}
Hu, J.; Liu, D.; Fu, N.; and Dong, R. 2024.
\newblock Realistic material property prediction using domain adaptation based machine learning.
\newblock \emph{Digital Discovery}, 3(2): 300--312.

\bibitem[{Jha et~al.(2019)Jha, Choudhary, Tavazza, Liao, Choudhary, Campbell, and Agrawal}]{jha2019enhancing}
Jha, D.; Choudhary, K.; Tavazza, F.; Liao, W.-k.; Choudhary, A.; Campbell, C.; and Agrawal, A. 2019.
\newblock Enhancing materials property prediction by leveraging computational and experimental data using deep transfer learning.
\newblock \emph{Nature communications}, 10(1): 5316.

\bibitem[{Jha et~al.(2022)Jha, Gupta, Liao, Choudhary, and Agrawal}]{jha2022moving}
Jha, D.; Gupta, V.; Liao, W.-k.; Choudhary, A.; and Agrawal, A. 2022.
\newblock Moving closer to experimental level materials property prediction using AI.
\newblock \emph{Scientific reports}, 12(1): 11953.

\bibitem[{Li et~al.(2024)Li, Fu, Omee, and Hu}]{li2024md}
Li, Q.; Fu, N.; Omee, S.~S.; and Hu, J. 2024.
\newblock MD-HIT: Machine learning for material property prediction with dataset redundancy control.
\newblock \emph{npj Computational Materials}, 10(1): 245.

\bibitem[{Li, Shan, and Shek(2022)}]{li2022machine}
Li, X.; Shan, G.; and Shek, C. 2022.
\newblock Machine learning prediction of magnetic properties of Fe-based metallic glasses considering glass forming ability.
\newblock \emph{Journal of Materials Science \& Technology}, 103: 113--120.

\bibitem[{{Martin Siron} et~al.(2024){Martin Siron}, {Inel Djafar}, {Lucile Ritchie}, {Etienne Du-Fayet}, {Amandine Rossello}, {Ali Ramlaoui}, {Leandro von Werra}, {Thomas Wolf}, and {Alexandre Duval}}]{lematerial_2024}
{Martin Siron}; {Inel Djafar}; {Lucile Ritchie}; {Etienne Du-Fayet}; {Amandine Rossello}; {Ali Ramlaoui}; {Leandro von Werra}; {Thomas Wolf}; and {Alexandre Duval}. 2024.
\newblock LeMat-BulkUnique Dataset.

\bibitem[{Pakala et~al.(2024)Pakala, Graw, Ahn, Dinh, Mahin, Tsotras, Chen, and Papalexakis}]{10825934}
Pakala, S.; Graw, B.; Ahn, D.; Dinh, T.; Mahin, M.~T.; Tsotras, V.; Chen, J.; and Papalexakis, E.~E. 2024.
\newblock Automating Data Science Pipelines with Tensor Completion.
\newblock In \emph{2024 IEEE International Conference on Big Data (BigData)}, 1075--1084.

\bibitem[{Paszke et~al.(2019)Paszke, Gross, Massa, Lerer, Bradbury, Chanan, Killeen, Lin, Gimelshein, Antiga et~al.}]{paszke2019pytorch}
Paszke, A.; Gross, S.; Massa, F.; Lerer, A.; Bradbury, J.; Chanan, G.; Killeen, T.; Lin, Z.; Gimelshein, N.; Antiga, L.; et~al. 2019.
\newblock Pytorch: An imperative style, high-performance deep learning library.
\newblock \emph{Advances in neural information processing systems}, 32.

\bibitem[{Pedregosa et~al.(2011)Pedregosa, Varoquaux, Gramfort, Michel, Thirion, Grisel, Blondel, Prettenhofer, Weiss, Dubourg, Vanderplas, Passos, Cournapeau, Brucher, Perrot, and Duchesnay}]{scikit-learn}
Pedregosa, F.; Varoquaux, G.; Gramfort, A.; Michel, V.; Thirion, B.; Grisel, O.; Blondel, M.; Prettenhofer, P.; Weiss, R.; Dubourg, V.; Vanderplas, J.; Passos, A.; Cournapeau, D.; Brucher, M.; Perrot, M.; and Duchesnay, E. 2011.
\newblock Scikit-learn: Machine Learning in {P}ython.
\newblock \emph{Journal of Machine Learning Research}, 12: 2825--2830.

\bibitem[{Sidiropoulos et~al.(2016)Sidiropoulos, De~Lathauwer, Fu, Huang, Papalexakis, and Faloutsos}]{sidiropoulos2016tensor}
Sidiropoulos, N.~D.; De~Lathauwer, L.; Fu, X.; Huang, K.; Papalexakis, E.~E.; and Faloutsos, C. 2016.
\newblock Tensor Decomposition for Signal Processing and Machine Learning.
\newblock \emph{IEEE Signal Processing Magazine}.

\bibitem[{Wang et~al.(2020)Wang, Tian, Kirk, Laris, Ross~Jr, Noebe, Keylin, and Arr{\'o}yave}]{wang2020accelerated}
Wang, Y.; Tian, Y.; Kirk, T.; Laris, O.; Ross~Jr, J.~H.; Noebe, R.~D.; Keylin, V.; and Arr{\'o}yave, R. 2020.
\newblock Accelerated design of Fe-based soft magnetic materials using machine learning and stochastic optimization.
\newblock \emph{Acta Materialia}, 194: 144--155.

\bibitem[{Xie and Grossman(2018)}]{xie2018crystal}
Xie, T.; and Grossman, J.~C. 2018.
\newblock Crystal graph convolutional neural networks for an accurate and interpretable prediction of material properties.
\newblock \emph{Physical review letters}, 120(14): 145301.

\bibitem[{Xin et~al.(2021)Xin, Siriwardane, Song, Zhao, Louis, Nasiri, and Hu}]{xin2021active}
Xin, R.; Siriwardane, E.~M.; Song, Y.; Zhao, Y.; Louis, S.-Y.; Nasiri, A.; and Hu, J. 2021.
\newblock Active-learning-based generative design for the discovery of wide-band-gap materials.
\newblock \emph{The Journal of Physical Chemistry C}, 125(29): 16118--16128.

\end{thebibliography}
